\documentclass[runningheads]{llncs}
\usepackage{graphicx}
\usepackage{booktabs}
\usepackage{hyperref}
\usepackage{color}
\usepackage{subfigure}
\usepackage{bbm}
\usepackage{amsmath,amssymb,amsfonts}

\begin{document}
\title{Learning-based Bone Quality Classification Method for Spinal Metastasis}


%
%

\author{Shiqi Peng\inst{1} 
\and Bolin Lai\inst{1} 
\and Guangyu Yao \inst{2}
\and Xiaoyun Zhang \inst{1}
\and Ya Zhang \inst{1}
\and Yan-Feng Wang \inst{1}
\and Hui Zhao \inst{2}
}

%

\institute{Cooperative Medianet Innovation Center, Shanghai, 200240, PRC \and
Shanghai Jiao Tong University Affiliated Sixth People's Hospital, 200233, PRC \\
\email{pengshiqi@sjtu.edu.cn, lai.b.bryan@gmail.com, ygy504187803@126.com, \{xiaoyun.zhang, ya\_zhang, wangyanfeng, zhao-hui\}@sjtu.edu.cn}}


\maketitle              

\vspace{-0.5cm}
\begin{abstract}
Spinal metastasis is the most common disease in bone metastasis and may cause pain, instability and neurological injuries.
Early detection of spinal metastasis is critical for accurate staging and optimal treatment.
The diagnosis is usually facilitated with Computed Tomography (CT) scans, which requires considerable efforts from well-trained radiologists. 
In this paper, we explore a learning-based automatic bone quality classification method for spinal metastasis based on CT images. 
We simultaneously take the posterolateral spine involvement classification task into account, and employ multi-task learning (MTL) technique to improve the performance. 
MTL acts as a form of inductive bias which helps the model generalize better on each task by sharing representations between related tasks. 
Based on the prior knowledge that the mixed type can be viewed as both blastic and lytic,
we model the task of bone quality classification as two binary classification sub-tasks, \textit{i.e.}, whether blastic and whether lytic, and leverage a multiple layer perceptron to combine their predictions.
In order to make the model more robust and generalize better, self-paced learning is adopted to gradually involve from easy to more complex samples into the training process.
The proposed learning-based method is evaluated on a proprietary spinal metastasis CT dataset. At slice level, our method significantly outperforms an 121-layer DenseNet classifier in sensitivities by $+12.54\%$, $+7.23\%$ and $+29.06\%$ for blastic, mixed and lytic lesions, respectively, meanwhile $+12.33\%$, $+23.21\%$ and $+34.25\%$ at vertebrae level. 

\end{abstract}

\vspace{-0.75cm}

\section{Introduction}
\label{introduction}

\vspace{-0.25cm}

Metastasis is the spread of cancer from one part of the body to another. Approximately two-thirds of patients with cancer will develop bone metastasis.
The spine is the most common site of bone metastasis. A spinal metastasis may cause pain, instability and neurological injuries. Thus, the early diagnosis of spinal metastasis is crucial to change the patients' prognosis and improve the clinical outcome.

In 2017, the Spinal Instability Neoplastic Score (SINS)~\cite{fox2017spinal} was developed for assessing patients with spinal neoplasia. It acts as a prognostic tool for surgical decision making.
Among the six components in the SINS system, both bone quality and posterolateral involvement of spinal elements can be diagnosed from the axial view. Thus we are motivated to leverage Multi-task Learning (MTL) technique to address these two issues simultaneously. 
Recently, more and more MTL methods are applied to the field of medical imaging. A multi-task residual fully convolutional network was proposed for the pelvic Magnetic Resonance Image (MRI) segmentation task, employing three regression tasks to provide more information for helping optimize the segmentation task~\cite{feng2018semi}. A multi-task convolutional neural network was utilized to automatically predict radiological scores in spinal MRIs~\cite{jamaludin2016spinenet}.
MTL can be viewed as a form of inductive transfer. Inductive transfer can help improve a model by introducing an inductive bias, which causes a model to prefer some hypotheses over others and generally leads to solutions that generalize better. 

\begin{figure}[tbp]
\centering
\label{fig1}

\subfigure[Normal.]{
\begin{minipage}[t]{0.2\linewidth}
\centering
\includegraphics[width=0.9\linewidth]{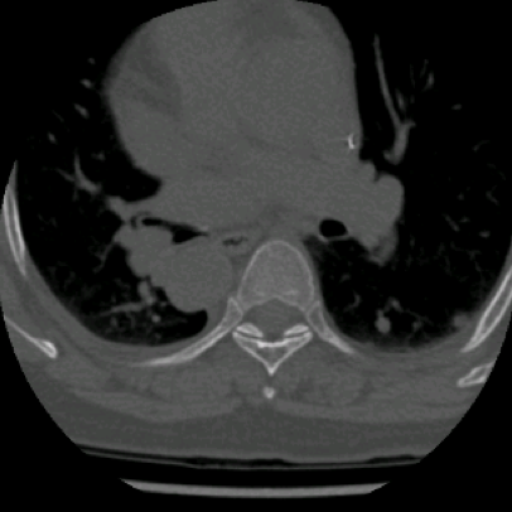}
\end{minipage}%
}%
\subfigure[Blastic.]{
\begin{minipage}[t]{0.2\linewidth}
\centering
\includegraphics[width=0.9\linewidth]{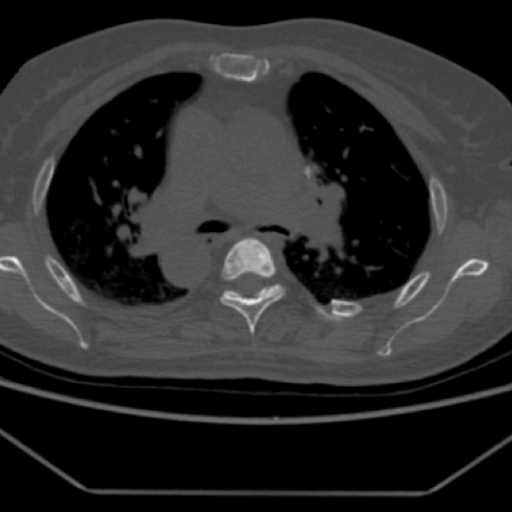}
\end{minipage}%
}%
\subfigure[Mixed.]{
\begin{minipage}[t]{0.2\linewidth}
\centering
\includegraphics[width=0.9\linewidth]{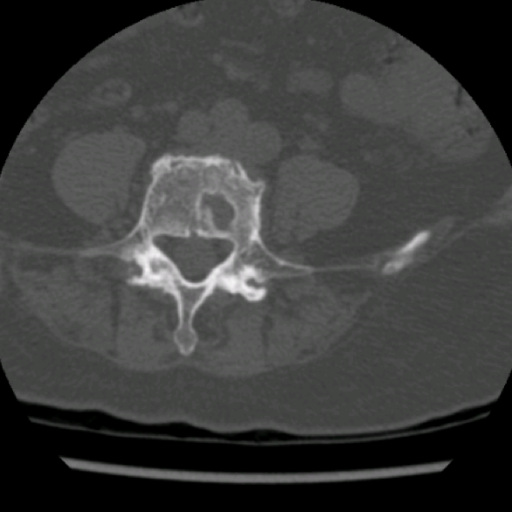}
\end{minipage}
}%
\subfigure[Lytic.]{
\begin{minipage}[t]{0.2\linewidth}
\centering
\includegraphics[width=0.9\linewidth]{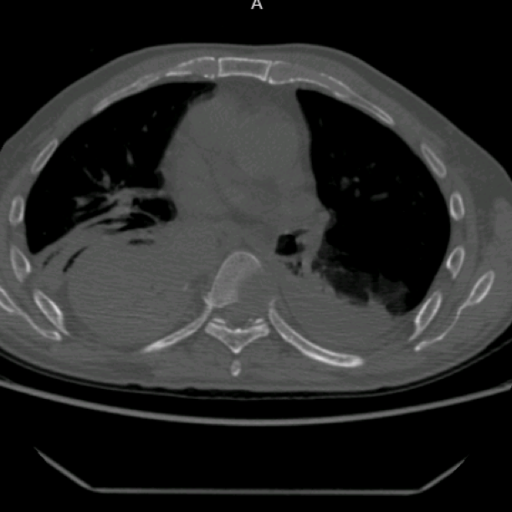}
\end{minipage}
}%

\centering
\vspace{-0.25cm}
\caption{Typical examples for bone quality of spinal metastasis.}
\vspace{-0.5cm}
\end{figure}

In the SINS system, metastatic bone quality is divided into three types (blastic, mixed and lytic, as shown in Fig. \ref{fig1}). 
Different types indicate different clinical outcomes and therapies. 
Previous work on bone lesion quality classification was primarily a binary classification of benign and malignant~\cite{kumar2016classification}.
To the best of our knowledge, we are the first to use deep neural networks to achieve automatic four-category bone quality classification. 
However, satisfactory results for this problem can not be achieved by simply using a deep neural network because (1) training data is limited, (2) the classification cues are fine-grained and (3) it is hard to learn the decision boundaries between categories. 
We then target to address these three issues. 
We handle this problem from 2D axial slices. Thus hundreds of training images can be collected from each patient. 
For an axial image, we adopt a threshold extraction method introduced in~\cite{guan2019deep} to increase SNR based on HU values and concentrate on the bone areas instead of soft tissues.
Then we model the task of bone quality classification as two binary classification sub-tasks, \textit{i.e.}, whether blastic and whether lytic.
In order to make the model more robust and generalize better, self-paced learning~\cite{kumar2010self} is adopted to gradually involve from easy to more complex samples into the learning process. 
Finally, the predictions of sub-tasks are combined to obtain the four-category predictions, and the slice predictions are merged to get the vertebrae predictions by a voting method.

\setlength{\belowcaptionskip}{-0.5cm}

The proposed method is evaluated on a proprietary spinal metastasis CT dataset. A four-category $121$-layer DenseNet classifier~\cite{huang2017densely} is selected as our baseline. 
At slice level, our method achieves an improvement in sensitivity of $+12.54\%$, $+7.23\%$ and $+29.06\%$ for blastic, mixed and lytic lesions, respectively, meanwhile an improvement of $+12.33\%$, $+23.21\%$ and $+34.25\%$ at vertebrae level. 
More importantly, our method is expected to assist radiologists in practical diagnosis.

\vspace{-0.5cm}

\section{Proposed Method}
\label{method}

\vspace{-0.25cm}
\subsection{Framework}
\label{framework}
\begin{figure}[tbp]
\vspace{-0.25cm}
\centerline{\includegraphics[width=1.0\linewidth]{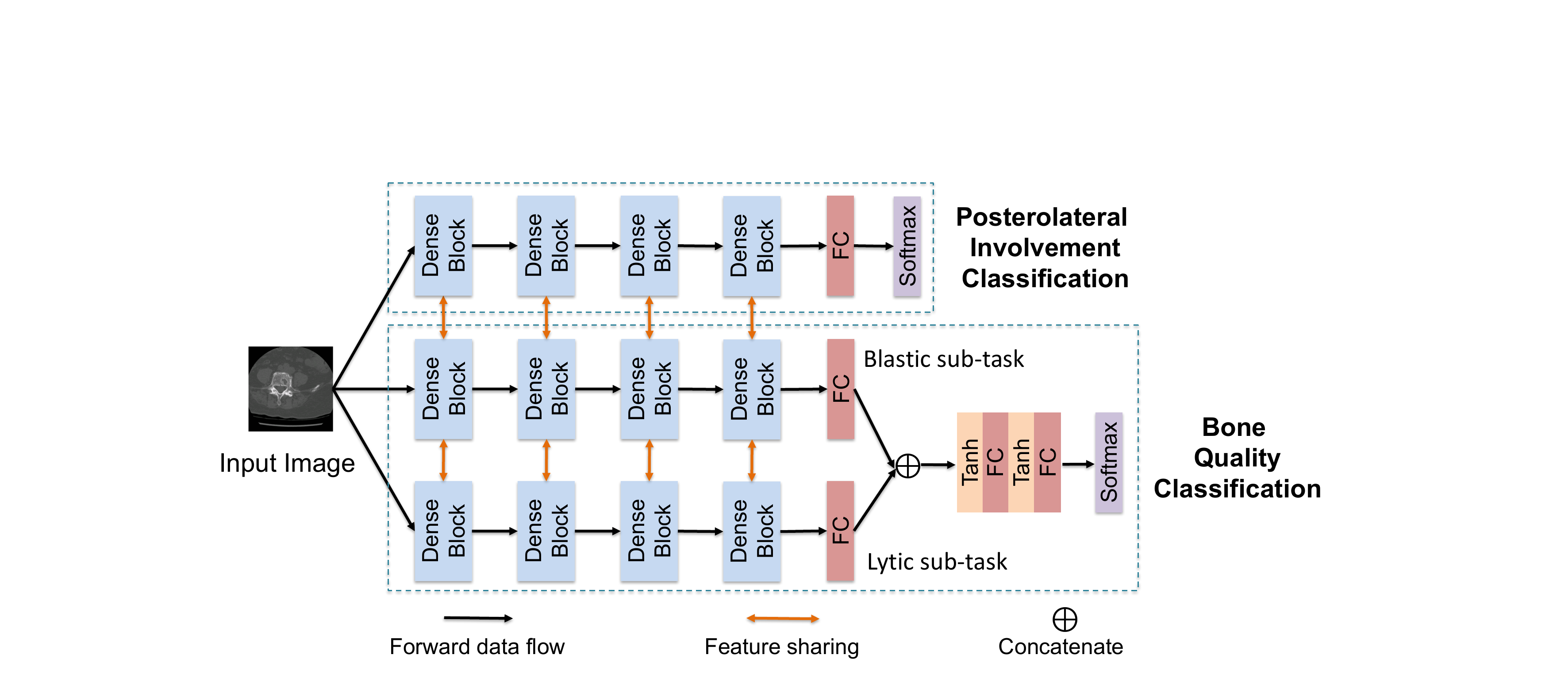}}
\caption{ The proposed multi-task learning framework. }
\label{fig2}
\end{figure}

\vspace{-0.25cm}
The framework of our proposed learning-based method is depicted in Fig. \ref{fig2}. Two tasks, bone quality classification and posterolateral involvement classification, are learned at the same time. Based on the prior knowledge that the mixed type can be viewed as both blastic and lytic, we model the task of bone quality classification as two sub-tasks, whether blastic and whether lytic. Similarly, the labels of training data can also be decomposed in the same way.


Now we implicitly have three tasks to learn and MTL is suitable for such situation.  
The premise of using MTL is that the tasks should be relevant so that they can be learned jointly. 
In our case, all these tasks have to extract features from the axial view which can be shared with each other.
Hence these tasks are strongly relevant and MTL technique is appropriate.
In order to make more use of the low-level features, DenseNet is selected as feature extractor for its dense skip connections. 
For each task, we use four dense blocks with $6, 12, 24, 16$ dense layers respectively. And the growth rate is $32$. Between each two adjacent dense blocks, a transition layer is employed for feature map downsampling. 

After obtaining the logits of two sub-tasks of bone quality, and passing them through two softmax layers, we can get the probabilities of an image being blastic and lytic respectively. 
For the final diagnosis, we have to combine these two predictions together. 
The general approach is to use a threshold to obtain the binary predictions and combine them. However, this approach causes information loss due to the thresholding operation.
We are expected to get the final prediction directly from the logits.
In order to learn such a mapping function,
we concatenate the logits from two sub-tasks and then pass them through a multiple layer perceptron with two linear layers and tanh activation function. 
Passing the embedded vectors through a softmax layer, the final  probabilities for four categories can be obtained.

\vspace{-0.5cm}
\subsubsection{Feature Sharing:} 
The feature sharing operation is performed at the end of each dense block. 
Feature sharing can be formulated as $[\tilde{x}_A, \tilde{x}_B] = f([x_A, x_B])$, 
where function $f$ is the feature sharing method, $A$ and $B$ are two relevant tasks, $x$ and $\tilde{x}$ are input and output features respectively.
In the context of deep learning, shared representation is typically done with either \textit{hard sharing} or \textit{soft sharing} of hidden layers. Hard sharing is the most commonly used approach, which is generally applied by sharing the hidden layers between all tasks while keeping several task-specific output layers. 
On the other hand in soft sharing, each task has its own model with its own parameters. The distance between the feature maps of the models is then regularized in order to encourage the feature maps to be similar.
In Sect. \ref{ablation_study}, we compare the performances of these two different kinds of shared representations and the experimental results demonstrate that hard sharing is the better choice for our problem.

\vspace{-0.5cm}
\subsubsection{Loss Function:}
The overall loss can be defined as 
\begin{equation}
\mathcal{L}_{overall} = \lambda_1 * \mathcal{L}_{Blastic} + \lambda_2 * \mathcal{L}_{Lytic} + \lambda_3 * \mathcal{L}_{BQ} + \mathcal{L}_{PI}
\label{eq1}
\end{equation}
where $\lambda_1$, $\lambda_2$ and $\lambda_3$ weight the relative contributions, \textit{BQ} and \textit{PI} denote for bone quality and posterolateral involvement respectively. All four loss function are cross entropy loss with the form $\mathcal{L} = -\sum_{i=1}^n y_i\log \hat{y}_i$, where $y_i$ is the ground truth and $\hat{y}_i$ is the prediction. In the experiments, we value the bone quality classification task more therefore $\lambda_1$, $\lambda_2$ and $\lambda_3$ are all set to be $2$. 

\vspace{-0.5cm}
\subsection{Self-paced Learning}
\vspace{-0.25cm}
During the training process, there exist many complex training examples disturbing the model optimization, such as sacral images and mis-labeled samples. To address this issue, we leverage self-paced learning (SPL) to gradually learn from easy to complex examples.

Formally, given a training set $\mathcal{D}=\{ (\mathbf{x}_i, y_i )\}_{i=1}^{n}$. Let $f(\cdot, \mathbf{w})$ denote the learned model and $\mathbf{w}$ be the model parameters. $L(y_i, f(\mathbf{x}_i, \mathbf{w}))$ is the loss function of $i$-th sample. The objective is to jointly learn the model parameter $\mathbf{w}$ and the latent weight variable $\mathbf{v}=[v_1, v_2, \dots, v_n]^\mathrm{T}$ by minimizing:
\begin{equation}
    \min_{\mathbf{w}, \mathbf{v} \in [0,1]^n}
    \mathbb{E}(\mathbf{w}, \mathbf{v}; \lambda)=\sum_{i=1}^{n} {v}_{i} L(y_i, f(\mathbf{x}_i, \mathbf{w})) - \lambda \sum_{i=1}^n v_i,
\label{eq2}
\end{equation}
where $\lambda$ is a penalty parameter that controls the learning pace. With the fixed $\mathbf{w}$, the global optimum $\mathbf{v^*}=[v_1^*, v_2^*, \dots, v_n^*]^\mathrm{T}$ can be calculated by:
\begin{equation}
v_i^* = \left\{
\begin{aligned}
1, & \quad L(y_i, f(\mathbf{x}_i, \mathbf{w})) < \lambda, \\
0, & \quad \text{otherwise.}   
\end{aligned}
\right.
\label{eq3}
\end{equation}
When alternatively updating $\mathbf{v}$ and $\mathbf{w}$, a sample with loss smaller than $\lambda$ is taken as an "easy" sample, and will be selected in training, or otherwise unselected. The parameter $\lambda$ controls the pace at which the model learns new samples.

\vspace{-0.5cm}
\subsection{Implementation Details}
\vspace{-0.25cm}
The usually used one-hot encoding for target label of cross entropy function encourages the output scores dramatically distinctive, which potentially leads to overfitting. Therefore the label smoothing technique~\cite{szegedy2016rethinking} is leveraged to address this issue. 
Besides, transfer learning technique is applied since the model is quite large but the training set is relatively small. We adopt weights for convolutional layers pre-trained on ImageNet, which help the model converge faster and achieve better results. In order to reduce the numerical instability and generalize better, the learning rate warmup strategy~\cite{xie2018bag} is also taken into consideration. These methods are analyzed in detail in Sect. \ref{ablation_study}.

\vspace{-0.5cm}
\section{Experimental Results}
\label{experiments}

\vspace{-0.25cm}
\subsection{Dataset}
\vspace{-0.25cm}
We evaluate the proposed method on a proprietary spinal metastasis dataset. This dataset contains $800$ CT scans come from patients with metastasis. The scans are reconstructed to in-plane resolution between $0.234$mm and $2.0$mm, and slice thickness between $0.314$mm and $5.0$mm. These CT scans cover four kinds of the spine, including cervical, thoracic, lumbar and sacral vertebraes. 
The reference labels are jointly annotated by three senior radiologists. 
Sub-sampling is used for data imbalance problem. The slices of each category for training are $9875$(normal), $9256$(blastic), $5846$(mixed) and $7164$(lytic) respectively, while $24133$, $698$, $642$ and $607$ for testing. 

\vspace{-0.5cm}
\subsection{Metrics}
\vspace{-0.25cm}
We use sensitivity(SE) and specificity(SP) as metrics to evaluate the classification performances.
For the bone quality classification, four categories are denoted as subscript N(normal), B(blastic), M(mixed) and L(lytic) in Tab. \ref{tab2}, Tab. \ref{tab4} and Tab. \ref{tab5}. 
For the auxiliary posterolateral involvement classification, three categories are denoted as subscript N(normal), U(unilateral) and B(Bilateral) in Tab. \ref{tab3}. 

\vspace{-0.5cm}
\subsection{Experiments}
\vspace{-0.25cm}
All the experiments are implemented in PyTorch\footnote{https://pytorch.org/}. 
SGD optimizer is used to optimize parameters with a learning rate and momentum of 0.001 and 0.9, respectively. We train the model for 50 epochs on an NVIDIA 1080 Ti GPU with 12GB memory. 
All the experiments are repeated five times and the averaged results are reported thus random errors are reduced. 

In general, the vertebraes only take up a relatively small part of the CT image. Therefore the classification cues are fine-grained and it is hard for model to concentrate on the bone areas. Based on the prior knowledge that the HU values of bones are relatively high than soft tissues, we adopt a threshold extraction method introduced in~\cite{guan2019deep} to extract the area of bone and crop the image from $512 \times 512$ to $224 \times 224$. 

\vspace{-0.5cm}
\subsection{Ablation Studies}
\label{ablation_study}

\begin{table*}[tbp]
\caption{Bone Quality Classification results at slice level.}
\begin{center}
\begin{tabular*}{\hsize}{@{}@{\extracolsep{\fill}}ccccccccc@{}}
\hline
                 & $SE_N$   & $SP_N$  & $SE_B$  & $SP_B$   & $SE_M$  & $SP_M$   & $SE_L$  & $SP_L$    \\
\hline \hline
\textbf{Single-task}         & \textbf{94.65} & 76.77          & 71.56          & \textbf{97.31} & 38.91          & 98.50 & 41.15          & \textbf{97.16}  \\
\textbf{Soft-sharing}   & 85.18 & 86.09          & 79.03          & 93.49 & 38.36          & 98.28 & 58.66          & 92.11  \\
\textbf{Hard-sharing-1} & 85.03 & 85.44          & 76.50          & 94.26 & 36.81          & 98.23 & 51.65          & 91.17  \\
\textbf{Hard-sharing-2} & 92.48 & 87.13 & 80.00 & 96.53 & 41.80 & 98.68 & 65.02 & 95.78  \\
\textbf{Hard-sharing-PI}      & 92.63 & \textbf{89.42}          & 80.23          & 96.69 & 41.67          & 98.80 & 66.84          & 95.51  \\
\textbf{Hard-sharing-MLP}   & 93.66 & 85.70 & 81.23 & 96.84 & 44.03 & \textbf{98.94} & 66.88 & 96.57  \\
\textbf{Final model}  & 92.43 & 88.88   & \textbf{84.10} & 96.08 & \textbf{46.14} & 98.93 & \textbf{70.21} & 96.15  \\
\hline   
\end{tabular*} 
\label{tab2}
\end{center}
\end{table*}

\vspace{-0.25cm}

\subsubsection{Feature sharing method:}
We first discuss the feature sharing method for the blastic and lytic classification tasks.
The forementioned soft feature sharing method and hard feature sharing method are investigated. The \textbf{single task} four-category classification is selected as baseline.
For soft sharing, we leverage the $\ell_2$ norm to regularize the feature maps after each dense block to be similar (denoted as \textbf{Soft-sharing}). In this situation, the overall loss function is defined as 
\begin{equation}
\mathcal{L}_{soft} = \mathcal{L}_{Blastic} + \mathcal{L}_{Lytic} + \sum_{i=1}^4 \lambda_i \| f_{Blastic}^i - f_{Lytic}^i \|_F^2
\label{eq4}
\end{equation}
where $f^i$ is the feature map of $i$th dense block, $\|\cdot\|_F$ is the Frobenius Norm and $\lambda_i$ are the weights.
For hard sharing, we compare the effect of sharing three dense blocks (denoted as \textbf{Hard-sharing-1}) and sharing four dense blocks (denoted as \textbf{Hard-sharing-2}). 
The results in Tab. \ref{tab2} shows that hard sharing all four dense blocks is the best shared representation for our tasks.

\begin{table}[tbp]
\caption{Posterolateral Involvement Classification results at slice level.}
\begin{center}
\begin{tabular*}{\hsize}{@{}@{\extracolsep{\fill}}ccccccc@{}}
\hline
                & $SE_N$   & $SP_N$  & $SE_U$  & $SP_U$   & $SE_B$  & $SP_B$    \\
\hline \hline
\textbf{Single task}       & \textbf{97.26} & 59.43          & 31.63          & \textbf{98.42} & 49.61          & \textbf{97.88}   \\
\textbf{MTL(ours)}         & 96.98 & \textbf{63.85} & \textbf{39.32} & 98.28 & \textbf{53.63} & 97.84   \\
\hline
\end{tabular*} 
\label{tab3}
\end{center}
\vspace{-0.5cm}
\end{table}

\vspace{-0.5cm}
\subsubsection{Posterolateral Involvement:}
We then explore the contribution of taking posterolateral involvement (PI) into account. 
The same hard sharing approach is adopted for PI and bone quality task.
The experimental results for bone quality classification with PI is shown in Tab. \ref{tab2} denoted as \textbf{Hard-sharing-PI}. Compared with the results without it (\textbf{Hard-sharing-2}), $SP_N$, $SE_B$ and $SE_L$ are all improved, while $SE_M$ remains basically unchanged. 

As for the PI task, experimental results are displayed in Tab. \ref{tab3}. 
The multi-task learning method is compared with the single task learning method by a 121-layer DenseNet classifier. It can be observed that $SP_N$, $SE_U$ and $SE_B$ have been greatly improved while other metrics have remained basically unchanged.
The above results show that joint learning bone quality and PI classification allows each task to generalize better.

\vspace{-0.5cm}
\subsubsection{Bone Quality:}
As for the label combination of the sub-tasks in bone quality classification, we investigate the performance improvement brought by MLP. Based on the hard sharing method, we adopt a two-layer perceptron with $10$ hidden units and tanh activation function. The sensitivities of three kinds of lesions are all promoted in Tab. \ref{tab2} (denoted as \textbf{Hard-sharing-MLP}). 


\vspace{-0.5cm}
\subsubsection{SPL and Implementations:}
\begin{table*}[tbp]
\caption{Bone Quality Classification results with learning methods at slice level. SPL, LS and LRS denote Self-Paced Learning, Label Smoothing and LR Scheduler, respectively. }
\begin{center}
\begin{tabular*}{\hsize}{@{}@{\extracolsep{\fill}}ccc|cccccccc@{}}
\hline
 SPL \quad &  LS \quad &  LRS \quad    & $SE_N$   & $SP_N$  & $SE_B$  & $SP_B$   & $SE_M$  & $SP_M$   & $SE_L$  & $SP_L$    \\
\hline \hline
 \quad & \quad &  \quad    & 92.48 & 87.13          & 80.00          & \textbf{96.53} & 41.80          & 98.68 & 65.02          & 95.78 \\
 \checkmark & \quad &   \quad      & 91.73 & 88.49          & 81.86          & 96.18 & 41.75          & 98.83 & 66.93          & 95.26 \\
 \checkmark & \checkmark &    \quad       & 91.87 & \textbf{89.14} & 82.26          & 96.45 & 44.24          & 98.80 & \textbf{68.74} & 95.22 \\
 \checkmark & \checkmark & \checkmark       & \textbf{92.84} & 87.22          & \textbf{84.11} & 95.93 & \textbf{48.37} & \textbf{99.12} & 66.31          & \textbf{96.59} \\
\hline   
\end{tabular*} 
\label{tab4}
\end{center}
\end{table*}

We then analyze the effect of self-paced learning (SPL) and other implementations in Tab. \ref{tab4}. 
LR scheduler contains learning rate warmup and cosine decay strategy. 
It is observed that these three modules can help the model generalization to a certain degree. When adding the LR scheduler to the previous model, $SE_B$ and $SE_M$ are improved while $SE_L$ is decreased, which can be considered as a tradeoff between mixed type and lytic type.

\vspace{-0.5cm}

\subsection{Final results}

\begin{table}[tbp]
\caption{Bone Quality Classification results at vertebrae level.}
\begin{center}
\begin{tabular*}{\hsize}{@{}@{\extracolsep{\fill}}ccccccccc@{}}
\hline
              & $SE_N$   & $SP_N$  & $SE_B$  & $SP_B$   & $SE_M$  & $SP_M$   & $SE_L$  & $SP_L$ \\
\hline \hline
\textbf{Single task}       & \textbf{96.71} & 72.89          & 70.77          & \textbf{97.49} & 36.23          & \textbf{97.86} & 38.39          & \textbf{97.71}  \\
\textbf{MTL(ours)}            & 88.47 & \textbf{87.71} & \textbf{83.10} & 95.47 & \textbf{59.44} & 97.52 & \textbf{72.64} & 94.02  \\
\hline
\end{tabular*} 
\label{tab5}
\end{center}
\vspace{-0.5cm}
\end{table}



\begin{figure}[tbp]
\centerline{\includegraphics[width=0.9\linewidth]{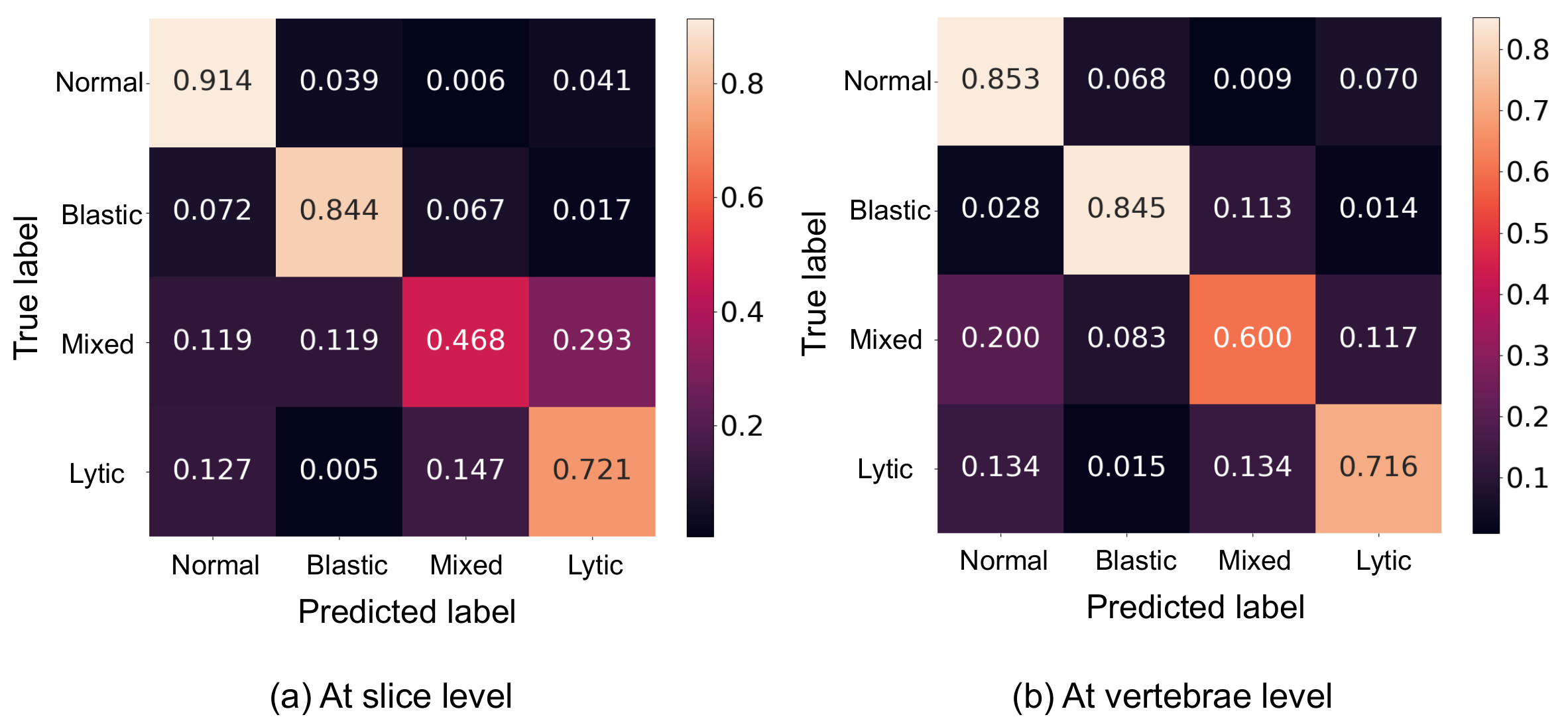}}
\caption{ Confusion matrices for bone quality at slice level and vertebrae level. }
\label{fig4}
\vspace{-1cm}
\end{figure}

\vspace{-0.25cm}

Finally, we combine all the components above to obtain the final model (denoted as \textbf{Final-model} in Tab. \ref{tab2}), which achieves the best performance on the $SE_B$, $SE_M$ and $SE_L$ outperforms baseline method by $+12.54\%$, $+7.23\%$ and $+29.06\%$, respectively. 
Fig. \ref{fig4} depicts the confusion matrices of the experiment for slices and vertebraes.
Tab. \ref{tab5} shows the results for vertebrae predictions.
The vertebraes are detected by an extra detection model. For each detected vertebrae, we use a voting method to obtain the vertebrae prediction from the slice predictions. 
If the maximum slice number of a kind of lesion is greater than a threshold, then the predict label of this vertebrae is determined as this kind of lesion. Otherwise the vertebrae is determined as normal. 
The proposed method achieves an improvement on $SP_N$, $SE_B$, $SE_M$ and $SE_L$ by $+14.82\%$, $+12.33\%$, $+23.21\%$ and $+34.25\%$, respectively. 
These improvements allow our approach to help doctors with laborious work in practice. 

\vspace{-0.5cm}

\section{Conclusion}
\label{conclusion}

\vspace{-0.3cm}

In this paper, we have explored an automatic learning-based method to classify the bone quality and posterolateral involvement of spinal metastasis. Multi-task learning helps both tasks generalize better by sharing representations. Besides, we model the task of bone quality classification as two sub-tasks and leverage a multiple layer perceptron to combine their predictions. Furthermore, in order to make the model more robust and generalize better, we adopt self-paced learning to gradually involve from easy to more complex samples into the training process. 
Adequate experiments on our proprietary dataset prove that the proposed method is effective. In the future, our method is expected to assist radiologists in practical diagnosis.

\vspace{-0.25cm}

%

\bibliographystyle{splncs04}
\bibliography{reference}

\end{document}